# Temporal Information and Event Markup Language

TIE-ML Markup Process and Schema Version 1.0

Damir Cavar, Billy Dickson, Ali Aljubailan, Soyoung Kim
Department of Linguistics
Indiana University
Bloomington, USA
email: {dcavar, dicksonb, aaaljuba, sk135}@iu.edu

*Abstract*—Temporal Information and Event Markup Language (TIE-ML) is a markup strategy and annotation schema to improve the productivity and accuracy of temporal and event related annotation of corpora to facilitate machine learning based model training. For the annotation of events, temporal sequencing, and durations, it is significantly simpler by providing an extremely reduced tag set for just temporal relations and event enumeration. In comparison to other standards, as for example the Time Markup Language (TimeML), it is much easier to use by dropping sophisticated formalisms, theoretical concepts, and annotation approaches. Annotations of corpora using TimeML can be mapped to TIE-ML with a loss, and TIE-ML annotations can be fully mapped to TimeML with certain under-specification.

*Keywords*-TIE-ML; Events; Time; Corpora; Machine Learning.

## I. INTRODUCTION

Natural languages use various means to express events and place them in time. Tense, aspect, mood, and modality make up the foundations of this process, and each language utilizes a certain combination of these tools to indicate temporal information.

Tense places an event on the timeline and is most often generated through either verbal inflection, or the use of auxiliaries. Comrie [3] groups tenses into two categories: absolute and relative. Absolute tenses refer to tenses that orient an event with respect to the present (e.g., simple present, simple past, simple future) whereas relative tenses are those that orient an event with respect to a different point in time (e.g., pluperfect, future perfect).

The internal temporal structure of an event can be broken into two categories, grammatical aspect and lexical aspect. Grammatical aspect refers to the marking of aspect through inflection or auxiliaries (e.g., English progressive *-ing*) whereas lexical aspect refers to the inherent temporal properties of a predicate. The original four categories for grouping predicates by their lexical aspectual properties as introduced by Vendler [27] are statives, activities, accomplishments, and achievements, each of which housing differing combinations of telic, dynamic, and durative properties of predicates.

Modality as defined in Portner [13] is "the linguistic phenomenon whereby grammar allows one to say things about, or on the basis of, situations which need not be real" (p. 8). Mood falls under this umbrella and indicates how a proposition expresses modality. Mood can be subdivided into two categories, verbal mood (indicatives and subjunctives) and sentence mood (declaratives, interrogatives, and imperatives). Modal auxiliaries like *may*, *might*, *can*, *should*, etc. express epistemic, deontic, and circumstantial modality.

Understanding these expressions and how they interact and complement each other is crucial toward developing a system for capturing time and event information in natural language. Developing corpora and data sets is essential for quantitative studies of distributional properties of temporal and event logic phenomena and expressions. It also allows us to develop machine learning based automatic annotation and processing of event sequencing and temporal aspect as for example duration.

### A. Event Sequencing

Sequencing of events and sub-events is an essential task that we address here. A general observation is that the presentation sequence of events in natural language discourse does not necessarily correspond to the temporal sequence that these events occur in. While in example (1) the presentation of sub-events corresponds to the underlying temporal sequence, in example (2) the presentation sequence does not match with the temporal sequencing.

(1) *Wash the veggies, chop them, and fry them.*
       1           2         3
(2) *Before you fry the veggies, wash, and chop them.*
             3           1       2

Observations suggest that sub-events occupy the same time slot or happen simultaneously, illustrated in (3). To address this aspect, these events or eventualities need to be indexed individually, with an independent time slot index. If integers could represent indices for events and sequence indices for time slots, then the sequencing would be generated with two tiers: the event index tier E, and the time slot tier T. In (3), it is successfully indicated that the event "John lived in Paris" and the event "Susan studied in Berlin" are overlapping in time slot 1.

(3) *John lived in Paris while Susan studied in Berlin.*
    E:    1                  2
    T:    1                  1

The reference to time slots in example (3) does not encode the information whether there is a total overlap, subsumption,

or partial overlap, but only the information that there is a time slot/span such that the two events 1 and 2 overlap during this time span 1. To simplify the annotation, we could think of events or eventualities expressed by predicates to be tuples indicating the event identifier and a corresponding time slot, e.g., in (3) it is the predicate "live" assigned (1,1), and the predicate "studied" assigned (2,1).

For independent reasons we restrict our exposition here to this simplified perspective of enumerating events or eventualities, referencing time slots or spans using integer identifiers.

*B. Tense*

Reichenbach [18] introduced a theory of tense that presents three time variables that include event time, reference time, and speech time. Event time refers to the time of an event. Reference time is the point of reference along the time axis that an event is related to. Speech time refers to the time of utterance of an event. In absolute tenses, such as English simple past, present and future, the reference time and event time are simultaneous, however in relative tenses such as the pluperfect and future perfect, the reference time and event time are separated. To represent this ordering relationship encoding an event's specific tense, the three variables E (event time), R (reference time), S (speech time) are ordered on the time axis. For example, the pluperfect would be represented by the event time E preceding the reference time R, and both preceding the speech time S, represented in the sequence E-R-S. Present tense could be described through the simultaneity or overlap of E, R, and S expressed as E,R,S in Reichenbach's model.

The relative sequence of R and S in Reichenbach's model reflects the tense categories *present* (S and R overlap), *past* (R precedes S), and *future* (S precedes R). The event is *anterior* to some reference time R, if E precedes R. It is *posterior*, if R precedes E, and it is *simple* if R and E overlap. This system derives tense category labels like Posterior Past or Posterior Future, that do not have correspondence in traditional tense labels like Future Perfect or Pluperfect.

Using the Reichenbach schema to encode tense of simple predicates allows us to correlate the encoding of tense with the sequencing of predicates. In addition, it is essential to encode scope relations between different predicates and tenses in complex clauses when considering assertions about events, whether they are claimed to be facts and true, or hypotheses about some future unfolding of events. Consider the examples in (4) and (5). While the predicate in (4) asserts that Apple merged with Alphabet, the predicate in (5) does not claim to be factual.

(4) *Apple merged with Alphabet.*
(5) *Apple will merge with Alphabet.*

If E precedes S, the event could be asserted to be factual, while S preceding E implies that the event is a hypothetical projection into the future.

The situation changes if a predicate with a tense as in (4) is in the scope of another predicate and specific tense. While example (6) is equivalent to (4) with respect to the embedded predicate, the matrix predicate and tense in (7) render the embedded predicate hypothetical.

(6) *Reuters reported that Apple merged with Alphabet.*
(7) *Reuters will report that Apple merged with Alphabet.*

It is essential to encode the tense of the individual predicates for the correct interpretation of the status of assertions. Syntactic scope relations between the predicates are necessary as well for the correct interpretation of embedded predicates.

The rest of the paper is structured as follows. In Section II, we present the overview of related work in the domain of temporal and event annotations of natural language corpora. In Section III, we describe the core properties of TIE-ML. In Section IV, we present our conclusions and the plan for future research related to TIE-ML. In Section V, we provide information about availability and open access to the TIE-ML standard and sample corpora.

## II. RELATED WORK

The demand for data sets and corpora with semantic annotation has grown over the last decades. One of the key types of information for Information Extraction (IE) systems to store, retrieve, and analyze is time and Temporal Expressions (TE).

The need to analyze and interpret event mentions in text sources or spoken language dialogues drive the necessity for deep understanding and models of event logic and temporal reasoning. Building temporally aware software systems can be significantly beneficial for Natural Language Processing (NLP) based information extraction applications, e.g., Question Answering Systems (QAS), Text Mining (TM) techniques, Document Summarization (DS) systems, Medical Documentation Systems (MDS), and other NLP applications such as event characterization and tracking and visualizing events on the timeline.

Accordingly, the automatic recognition to temporal elements in digital texts has recently turned out to be a vast area of research in the field of NLP; several activities and various initiatives were made attempting to develop representations for temporal information annotation in order to obtain more efficient information extraction.

This can account for the proliferation in research in this area, disseminated in theoretical bases and practical applications. The vast majority of work on annotating TEs, however, has been developed by three language technology evaluation programs: Message Understanding Conference (MUC) [34], the succeeding Translingual Information Detection, Extraction and Summarization (TIDES) [35], and The Automatic Content Extraction (ACE) [4]. All of these programs were held under the Defense Advanced Research Project Agency (DARPA) conference, sponsored by the U.S. government. In the remainder of this section, we briefly review the main existing schemes for annotating temporal information.

*A. MUCs*

MUCs are a series of language technology evaluation conferences "in which participating IE systems are rigorously evaluated" [1].

As previously mentioned, a considerable amount of work on IE has been cultivated by MUCs [21][29]. It is no surprise, then, that efforts on devising temporal annotation schemes appear to have begun within the DARPA MUCs [12][29]. More specifically, temporal information was defined as a targeted type for IE starting from MUC-6 and continuing in MUC-7. In MUC-6, one of the required subtasks for annotating Named Entity Recognition (NER) was to identify *absolute* time expressions in documents.

In MUC-7, the requirement of this subtask was extended to include *relative* time expressions [22]. It is worth noting that the distinction between absolute and relative time expressions was first made within MUCs guidelines [26]. However, none of the mentioned subtasks required the consideration of placing events in time or mapping temporal relations between events [22].

During MUC-7, the participating systems were required to extract the TIMEX elements, i.e., the TEs textual span, without being required to describe the inward semantic characteristics of the successfully deciphered TEs. The requirements then were to merely extract the coarse-grained type classification of each recognized TE. That is, the participating systems were required to determine whether a TIMEX represents a DATE or a TIME feature. Examples on such annotation requirement include the following:

- `"twelve o'clock noon"`
  `<TIMEX TYPE="TIME">twelve o'clock noon</TIMEX>`
- `"5 p.m. EST"`
  `<TIMEX TYPE="TIME">5 p.m. EST</TIMEX>`
- `"third quarter of 1991"`
  `<TIMEX TYPE="DATE">third quarter of 1991</TIMEX>`

*B. TIDES*

TIDES was a DARPA-sponsored research program on IE, specifying guidelines that were concerned with the specification and standardization of more detailed semantic representations of TEs than TIMEX had applied in the previous DARPA programs (see MUC 1998). However, TIDES guidelines maintain similarity to MUC's guidelines in handling TEs as separate targets for annotation and/or extraction. Additionally, those standards of TIDES were not aimed at the "hopelessly ambitious goal" of representing the entire varieties of TI expressed in natural language [7].

In the latest version of TIDES [7], *markable* expressions to be annotated must represent an appropriate *lexical trigger*. Also, a trigger must be able to be orientable on a timeline or at least be orientable to a relation to a time (past, present, future). Based on these determinations, lexical triggers that are reliable candidates of markable expressions are:

- nouns (*day*, *month*, *summer*, etc.)
- proper names (*Monday*, *January*, *New Year's Eve*)
- specialized time patterns (*8:00*, *12/2/00*, *1994*)
- adjectives (*recent*, *former*, *current*, *future*, *past*, *daily*, *monthly*, *biannual*, etc.)
- adverbs (*lately*, *hourly*, *daily*, *monthly*, etc.)
- noun or time adverb where adverbs that stem from an adjectival form of a trigger are also triggers. (*now*, *today*, *yesterday*, *tomorrow*, etc.)
- and numbers (*Sixties*, as in referring to the decade "the Sixties")

For temporal annotation format, TIDES developed a special SGML (Standard Generalized Markup Language) tag, i.e., TIMEX2, superseding MUC-7 TIMEX and extending its annotation. TIMEX2 offers a variety of features for more precise capturing of the actual meaning of a TE. TIMEX2 therefore is claimed to be most easily applicable to languages other than English, although all the cases defined and discussed in [5] are related to English.

*C. ACE*

The Automatic Content Extraction (ACE) is a program created by The National Institute of Standards and Technology (NIST) that is driven by and addresses issues identical to MUCs. ACE is a series of evaluation activities that require developing human language technologies capable of understanding natural language, thereby being automatically capable of detecting and extracting the key types of information existed in digital multimedia resources. One of these key types is events with associated entities and their temporal anchoring, which were added to ACE IE efforts in 2004 [4].

By collaboration, the Linguistic Data Consortium (LDC) at the University of Pennsylvania developed annotation guidelines, annotated corpora, and produced other linguistic resources to support the ACE program for research on IE. One of the primary ACE annotation tasks was Event Detection and Characterization (EDC).

In EDC, annotators identified and characterized five types of events in which EDT entities participated. Targeted types included Interaction, Movement, Transfer, Creation and Destruction events. Annotators tagged the textual mention or anchor for each event and categorized them by type and subtype. They further identified event arguments (agent, object, source, and target) and attributes (temporal, locative, instrument, purpose, etc.) according to a type-specific template. In later phases of ACE, annotators identified additional event types as well as characterized relations between events (see [33]).

### D. STAG

Sheffield Temporal Annotation Guidelines (STAG), analogous to the development of TIDES, is a TI annotation scheme that was created by Andrea Setzer for her PhD thesis [22][24]; Setzer's work is said to be the first annotation scheme ever to allow for all elements of TI [25]. In her framework, Setzer's objective was to annotate events, TEs, and their temporal relations. This framework is based on four primitive types: *events, states, times* and *relations*.

Event in STAG is intuitively defined as something that happens, must be anchorable in time map, and can be ongoing or conceptually instantaneous [23]. Based on this simple definition, in her scheme Setzer categorizes events into coarse-grained sets, including *occurrence, reporting, perception, attitude*, and *aspectual events*.

For time, instead of viewing times as having extents (intervals), or as being punctual (having a time point), STAG simply applies the notion of *time objects*. Time objects must be replaceable on a timeline and are either fictional or real [24]. Following the broad conventions of MUC's approach in labeling time, time objects in STAG are classified into two types, DATES and TIMES, where times are broadly described as being larger or smaller than a day.

Regarding temporal relations, STAG defines relations between events and other events, and events and times. The framework provided for temporal relations heavily depends on the works on temporal relations and temporal ontology conducted by Allen [2][24]. As a result, in providing a practical framework for temporal relations, the set of relations that connect events to times was reduced to merely five relations: *before, after, includes, included*, and *simultaneous*, the latter being vague to determine [28].

### E. TimeML

TimeML [15][17] is a metadata standard proposed for TI annotation, and it is currently the most conventional mark-up language for annotating events and temporal relations [8][16][17].

The framework of TimeML was created based on recommendations from the Time and Event Recognition for Question Answering Systems (TERQUAS) workshop in July 2002. TERQUAS feedback was given on how to enhance temporally aware NLP question answering systems (QAS) [9][11][12].

Pustejovsky and his colleagues proposed the TimeML specifications for annotating events and their temporal anchoring by amalgamating two of the previous TI annotation schemes: TIMEX2 [5][6][7] and STAG, along with other emerging schemes such as in Katz and Arosio [12].

Dissimilarly from the previous attempt at specifying event and time, TimeML separates the representation of event and temporal expressions from the anchoring or ordering dependencies that may exist in a given text.

There are four major structures specified in TimeML [15][17]: EVENT, TIMEX3, SIGNAL, and LINK. The tag <EVENT> is a cover term for the ontological notion of "events": situations that happen or occur, either punctually or as lasting for a period of time. Events in TimeML are broadly expressed by several linguistic formations, including *verbs, nominalizations, adjectives, predicative clauses*, and *prepositional phrases*.

The TIMEX3 tag, which is used for marking up explicit TEs, e.g., times and dates, is based on both the TIMEX [24] and TIDES TIMEX2 tag [6].

The use of signals is another feature of TimeML that was originally borrowed from Setzer's STAG then expanded in TimeML. The tag <SIGNAL> is used to annotate function words, i.e., indicators of temporal relations, such as temporal connectives (e.g., *while*), or temporal prepositions (e.g., *during*).

The fourth tag, <LINK>, said to be a key innovation for TimeML [15], comprises three types of link tags: TLINK, SLINK, and ALINK. The main task of the <LINK> tag is to encode relations between temporal elements in a text. TimeML proposes a set of 13 relations to indicate fine-grained distinctions between TEs and/or between TEs and events. Overall, the features that distinguish TimeML from other previous schemes below are:

- Extends the TIMEX2 attributes
- Introduces Temporal Functions to allow intentionally specified expressions: *three years ago, last month*
- Identifies signals determining interpretation of temporal expressions
  - Temporal prepositions: *for, during, on, at*;
  - Temporal Connectives: *before, after, while*.
- Identifies all classes of event expressions
  - Tensed verbs: *has left, was captured, will resign*
  - Stative adjectives and other modifiers; *sunken, stalled, on board*
  - Event nominals: *merger, Military Operation, Gulf War*
- Creates dependencies between events and times:
  - Anchoring: *John left on Monday*.
  - Orderings: *The party happened after midnight*.
  - Embeddings: *John said Mary left*.

### III. TIE-ML

While TimeML represents an approved, very detailed and precise annotation standard for events and temporal relations, it also introduces a high level of complexity for annotators. In our practical lab experience, the time and complexity to annotate basic data sets was prohibitively high. It required experts and well-trained linguistic annotators, and the productivity and quality control turned out to be costly. TIE-ML is a solution for a basic event sequencing corpus with Reichenbach style of tense annotation that reduces the annotation complexity and facilitates much faster output with less errors.

The TIE-ML annotation system is designed to improve the accuracy for annotators by simplifying the annotation task for time and event information. Speeding up the annotation by reducing the complexity of the effort for annotators will hopefully lead to larger data sets in shorter time, reducing costs and annotation errors.

At the same time, the goal of TIE-ML is to facilitate machine learning model development for event sequence annotation and event labeling. To experiment with automatic sequencing, very basic annotations are necessary, as for example a basic event annotation for the presentation sequence and the time sequence.

*A. Event Identification*

Events in the TIE-ML schema are individual predicates that are usually clauses. Each clause or independent predicate is given a numerical event identifier (`eventid`), shown in Figure 1, that serves both to mark relationships between events, as well as track the presentation order of events in text.

Since the temporal ordering of events does not necessarily coincide with the presentation order, tracking this information can provide insight in the intuition and motivations of an author or interlocutor for presenting events in a particular way.

```
<s> <c eventid="1">
    Danny watched the movie
    </c>
    <c eventid="2">
    and ate popcorn
    </c>. </s>
<s> <c eventid="3">
    Josh brought the pizza
    </c>. </s>
```

Figure 1. EventID

*B. Tense, Perfect, Progressive*

For each event, TIE-ML provides the possibility for the tense of the predicate, as well as the presence of perfect and progressive aspect to be explicitly annotated using the *tense* attribute, and Boolean *perfect* and *progressive* attributes as shown in Figure 2.

Progressive provides information on the internal temporal structure of the event, while tense and perfect aspect provide information on the location of the event in time and the point the event is oriented with relationship to.

```
<s> <c tense="PAST" perfect= "TRUE"
       progressive="TRUE">
    The patient had been experiencing
    stomach pain
    </c>. </s>
```

Figure 2. Tense, Perfect, Progressive

As described below, Reichenbach's time variables provide a more specific annotation that provides more information than the traditional tense labels.

Some languages utilize morpho-syntactic present tense to refer to future events. In Polish, for example, the use of present tense verb forms is compatible with adverbials that indicate future tense reference, as in (8).

(8) *Jutro     pracuję         od   9 do 5.*
    Tomorrow work-1st-sg-present from 9 to 5
    "Tomorrow I will work from 9 to 5."

The same is not possible for past tense adverbials, as in the ungrammatical example (9).

(9) *\*Wczoraj   pracuję         od   9 do 5.*
     Yesterday work-1st-sg-present from 9 to 5

The convention in TIE-ML is to encode the semantic temporal properties in such constructions, assuming that morphosyntactic and part-of-speech annotation tools will provide the lexical level annotation, indicating present tense at the lexical and syntactic level.

*C. Reichenbach Annotation Model*

In TIE-ML, Reichenbach's [18] time variables are annotated as E for "event time," R for "reference time," and S for "speech time."

In absolute tenses such as English simple past, present and future, R and E are simultaneous, however in relative tenses such as the pluperfect and future perfect, R and E are separated. To represent this ordering relationship, the three time variables are each assigned an integer value from -2 to 2. These values represent a simple relationship where a variable with a lower value occurs before those with higher values, and variables with equivalent values occur simultaneously.

For the simple past sentence in Figure 3, the event time and reference time are simultaneous in the past and given a value of -1 relative to the speech time given a value of 0.

```
<s> <c E="-1" R="-1" S="0">
    Danny watched the movie.
</c> </s>
```

Figure 3. Reichenbach Simple Past

For the pluperfect sentence in Figure 4, the event occurs in the past relative to the reference time, and the reference time occurs in the past relative to the speech time. The event time is given a value of -2, reference time a value of -1, and speech time a value of 0.

```
<s> <c E="-2" R="-1" S="0">
    Josh had watched the movie.
</c> </s>
```

Figure 4. Reichenbach Pluperfect

In addition to encoding the sequencing of E, R, and S, negative and positive numbers are assigned to the variables to indicate past, present, and future directly in the value. While relative ordering of the values is sufficient for the derivation of tense categories and traditional labels for tense in the Reichenbach model, a negative value indicates past, a positive value future, and 0 corresponds to present tense. We exploit this property in different approaches to corpus analysis and machine learning model training.

*D. Reference Time Anchor*

To be able to capture the concrete reference time for an event, we provide a designated attribute to capture concrete date or time point expressions that anchor R on the real time axis.

The reference tag as in Figure 5 marks explicitly mentioned time and dates of events in text or conversations. This value provides a concrete temporal anchor for R in the Reichenbach model and the TIE-ML schema.

```
<s> <c reference="264 BC">
    The First Punic War broke out on the
    island of Sicily in 264 BC.
</c> </s>
```

Figure 5. Reference

*E. Timeline Sequencing*

To capture the presentation time and the relative timeslot association of events, the TIE-ML schema provides a timeslot attribute representing in its value the relative ordering of events along the time axis.

In Figure 6, we observe *starting the oven* as the second presented event that is assigned a timeslot value of 1, while the first presented event *prepare the vegetables* is assigned a timeslot value of 2. The temporal connective *after* signals this shift in the temporal order of events.

```
<s> <c eventid="1" timeslot="2">
    Prepare the vegetables </c>
    <c eventid="2" timeslot="1">
    after starting the oven
</c>. </s>
```

Figure 6. Timeline Slots

## IV. CONCLUSION AND FUTURE WORK

The most important goal for TIE-ML was to define an annotation schema that facilitates quick and reliable annotation of events and temporal sequencing, and bootstraps larger corpora in shorter time. The priority was given to solving the alignment of presentation and timeline sequencing. At the same time, clause level scope effects on tense in complex sentences with multiple verbal predicates can be annotated without significant increase of annotation effort. Various other aspects of event and temporal logic have been postponed to future versions and specifications.

Important components that are not yet integrated in TIE-ML are for example duration of events, or continuity and sequencing properties of events.

Continuity or sequencing of events is displayed in the contrast between the two predicates in (10) and (11).

(10) *John was reading the book for two months.*
(11) *John was living in New York for ten years.*

While the "reading" event in (10), based on common sense, would be understood to be a sequence of discontinuous sub-events of "reading." Common sense dictates that "living" in (11) is understood to be continuous. In a future release TIE-ML might provide a simple annotation attribute to indicate continuity of events.

Duration aspects are discussed in more detail in the following sub-section.

*A. Durations*

An additional development to TIE-ML is to mark deeper internal properties of events in the form of durations. One aspect of durations to be accounted for are set relations between event reference times. Consider the following examples. (12) sets a reference time of Monday, while (13), (14), and (15) set reference times that are subsets of 'Monday'.

(12) Event 1: *The test took place on <u>Monday</u>.*
(13) Event 2: *In the <u>morning</u> the students ate breakfast.*
(14) Event 3: *In the <u>afternoon</u> the students arrived.*
(15) Event 4: *In the <u>evening</u> the test finished.*

To capture this relationship, TIE-ML could introduce a tier system whereby each tier represents different sizes of reference times. Figure 7 displays two such tiers, one consisting of days, and the other, times of day. Events would then be linked through combinations of timeslot annotations and markers denoting an event's superset.

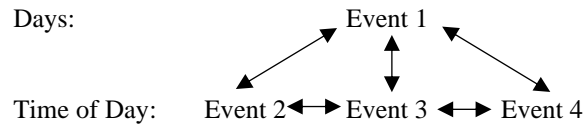

Figure 7. Duration Sets

Building further on capturing properties of durations, TIE-ML intends to incorporate an analysis of typical event duration allowing the execution of common-sense reasoning. For example, in (16), it certainly could be the case that Geoff very much does not like his vegetables, but this edge case aside, we can reason that an 'eating' event typically does not typically take 8 hours.

(16) *Geoff ate his dinner in 8 hours.*

We can also expand logicality prediction to the Reichenbach E, R, S variable values. In example (17) we observe a future tense event time in the past with respect to speech time, and a reference time in the future with respect to speech time. The simple future tense of this event calls for both reference time and speech time to have identical positive values. The inability to satisfy this requirement results in a logical incoherence.

(17) *#Yesterday I will go to the store.* E = 1, R = -1, S = 0

Various related aspects of complete or partial temporal overlap of events based on duration cannot be expressed in the current version of TIE-ML. Probabilities or estimates of duration for event and time related common-sense reasoning or models of expectation are not foreseen yet, but might likely enter a future version of TIE-ML.

*B. Cross-linguistic Features*

The current iteration of TIE-ML can be most effectively applied to tensed languages such as German in Figure 8, and the Semitic languages such as Arabic in Figure 9, and Hebrew in Figure 10.

```xml
<s> <c eventid="1" tense="FUT"
       reference="morgen"
       E="1" R="1" S="0">
    Morgen werde ich ein Buch lesen
    <!-- Tomorrow I will read a book -->
</c>. </s>
```

Figure 8. TIE-ML German

```xml
<s> <c eventid="1" timeslot="2"
       tense="sp" E="1" R="1" S="2">
           נתן נפגש עם רביקה </c>
    <c eventid="2" timeslot="1"
       tense="sp" E="1" R="1" S="2">
           לאחר שסיים לעבוד אחר הצהריים
<!-- Nathan met with Rebecca after he
finished working this afternoon -->
</c>. </s>
```

Figure 9. TIE-ML Hebrew

```xml
<s> <c eventid="1" timeslot="2"
       tense="sp" E="1" R="1" S="2">
           التقى أحمد بصديقه </c>
    <c eventid="2" timeslot="1"
       tense="sp" E="1" R="1" S="2">
           بعد أن زار والده الثلاثاء الماضي
<!-- Ahemd met with his friend after he
visited his father last Tuesday -->
</c>. </s>
```

Figure 10. TIE-ML Arabic

When expressing temporality in Semitic languages, ordinarily Arabic and Hebrew, there is a renewed controversy about whether tense is lacking in terms of grammatical expression, thereby making Semitic languages "aspect languages". Yet, even if this argument is conceded, not all aspectual dimensions are fully grammaticalized in Semitics [30][31]; that is, the main aspectual dimensions are Perfective (PFV) and Imperfective (IPFV), the latter being used to capture both simple and progressive situations. This is for the progressive aspect is not morphologically encoded thus not distinguished from habitual situation [32]. Rather, it is frequently indicated by other elements such as gerunds, adjectives, and adverbs. To capture these additional features, as well as the time and event information of languages that do not mark for tense requires additional annotations. Mandarin, for example, relies on a multitude of aspectual markers, temporal adverbials, and context to orient events. Mayan has a limited set of temporal adverbials and contains no temporal connectives, relying solely on aspect, mood, and context to order events.

Additional annotations of temporal features will enable TIE-ML to be applied to a wider range of typologically diverse languages. Some of these temporal features will be related to pragmatic cues and general world knowledge applied in general deduction or induction processes, often associated with probabilities or plausible guesses of temporal relations. While there is an obvious need for such annotation levels, in particular encoding of uncertainties and ambiguities, TIE-ML does not yet provide the means for such annotations.

The main motivation for defining the initial version of TIE-ML was to facilitate the development of cross-linguistic data sets with basic event and temporal logic annotation. The annotation process is much simpler than using other annotation standards and processes that certainly are far more sophisticated and powerful. The annotators have to understand only the three Reichenbach variables and event enumeration using event IDs. The training effort for annotators is clearly reduced in TIE-ML when compared to TimeML's rich tag and concept set. Future evaluations will provide more insight in the annotation error rate and output quantities.

Additionally, the simplified TIE-ML standard should be compatible with other annotation standards, provided core translation possibilities. Translation of TIE-ML annotations to other formats is straightforward, keeping in mind that TIE-ML provides certain levels of under-specification.

## V. AVAILABILITY AND OPEN ACCESS

Sample corpora and data sets are made available at the public GitHub repository "Temporal Information and Event Markup Language" (URL: https://github.com/dcavar/tieml). This repository also hosts conversion scripts and annotation samples for different languages.

The mapping of TIE-ML XML annotations from XML to different formats is straightforward. There are corpus samples of a proposed CoNLL format (URL: https://www.signll.org/conll/) mapping, as well as a JSON format annotation. The TIE-ML project will provide conversion scripts in the public GitHub repo for the unidirectional conversion to these formats and TimeML.

The Apache License Version 2.0 has been chosen as the appropriate license for the XML Schema, sample corpora, scripts, and documentation, facilitating commercial and non-commercial use.